\documentclass[twocolumn,10pt]{article}

\usepackage[utf8]{inputenc}
\usepackage[T1]{fontenc}
\usepackage{times}
\usepackage{amsmath,amssymb,amsfonts}
\usepackage{graphicx}
\usepackage{xcolor}
\usepackage{hyperref}
\usepackage{booktabs}
\usepackage{colortbl}
\usepackage{caption}
\usepackage{tikz}
\usepackage{pgfplots}
\pgfplotsset{compat=1.18}
\usepackage{geometry}
\usepackage{float}
\usepackage{enumitem}
\usepackage{multirow}
\usetikzlibrary{arrows.meta,shapes.geometric,positioning,calc,patterns,fit}

\definecolor{tableheader}{rgb}{0.2,0.4,0.6}
\definecolor{tier3color}{rgb}{0.8,0.9,1.0}
\definecolor{tier2color}{rgb}{0.7,0.85,0.7}
\definecolor{tier1color}{rgb}{1.0,0.95,0.8}
\definecolor{tier0color}{rgb}{1.0,0.9,0.9}
\definecolor{highlightgreen}{rgb}{0.9,1.0,0.9}
\definecolor{highlightyellow}{rgb}{1.0,1.0,0.85}
\definecolor{profileblue}{rgb}{0.1,0.3,0.5}
\definecolor{profilegreen}{rgb}{0.1,0.4,0.2}
\definecolor{profilepurple}{rgb}{0.3,0.1,0.4}

\hypersetup{
    colorlinks=true,
    linkcolor=blue,
    filecolor=magenta,      
    urlcolor=blue,
    citecolor=blue,
    pdftitle={SiliconHealth: Low-Cost Blockchain Healthcare Infrastructure},
    pdfauthor={Francisco Angulo de Lafuente, Seid Mehammed Abdu, Nirmal Tej Kumar},
}

\title{\textbf{SiliconHealth: A Complete Low-Cost Blockchain Healthcare Infrastructure for Resource-Constrained Regions Using Repurposed Bitcoin Mining ASICs}}

\author{
    \textbf{Francisco Angulo de Lafuente}$^{1}$, 
    \textbf{Seid Mehammed Abdu}$^{2}$, 
    \textbf{Nirmal Tej Kumar}$^{3}$\\[0.5em]
    {\small $^{1}$Independent Researcher, Madrid, Spain}\\
    {\small ORCID: \href{https://orcid.org/0009-0001-1634-7063}{0009-0001-1634-7063}}\\[0.3em]
    {\small $^{2}$Senior Lecturer, Department of Computer Science, Woldia University, Ethiopia}\\
    {\small Email: \href{mailto:seid.m@wldu.edu.et}{seid.m@wldu.edu.et}}\\[0.3em]
    {\small $^{3}$Consultant / Senior Staff Scientist, antE Inst (UTD), Dallas, TX, USA}\\
    {\small ORCID: \href{https://orcid.org/0000-0002-5443-7334}{0000-0002-5443-7334}}
}

\date{January 2026}

\begin{document}

\maketitle

\begin{abstract}
This paper presents SiliconHealth, a comprehensive blockchain-based healthcare infrastructure designed for resource-constrained regions, particularly sub-Saharan Africa. We demonstrate that obsolete Bitcoin mining Application-Specific Integrated Circuits (ASICs) can be repurposed to create a secure, low-cost, and energy-efficient medical records system. The proposed architecture employs a four-tier hierarchical network: regional hospitals using Antminer S19 Pro (90+ TH/s), urban health centers with Antminer S9 (14 TH/s), rural clinics equipped with Lucky Miner LV06 (500 GH/s, 13W), and mobile health points with portable ASIC devices. We introduce the Deterministic Hardware Fingerprinting (DHF) paradigm, which repurposes SHA-256 mining ASICs as cryptographic proof generators, achieving 100\% verification rate across 23 test proofs during 300-second validation sessions. The system incorporates Reed-Solomon LSB watermarking for medical image authentication with 30-40\% damage tolerance, semantic Retrieval-Augmented Generation (RAG) for intelligent medical record queries, and offline synchronization protocols for intermittent connectivity. Economic analysis demonstrates 96\% cost reduction compared to GPU-based alternatives, with total deployment cost of \$847 per rural clinic including 5-year solar power infrastructure. Validation experiments on Lucky Miner LV06 (BM1366 chip, 5nm) achieve 2.93 MH/W efficiency and confirm hardware universality. This work establishes a practical framework for deploying verifiable, tamper-proof electronic health records in regions where traditional healthcare IT infrastructure is economically unfeasible, potentially benefiting over 600 million people lacking access to basic health information systems.

\end{abstract}

%==============================================================================
\section{Introduction}
%==============================================================================

Sub-Saharan Africa faces a critical healthcare infrastructure crisis. According to the World Health Organization, the region has only 3\% of the world's health workers serving 24\% of the global disease burden~\cite{who2016}. Electronic health record (EHR) adoption remains extremely low---studies indicate that implementation is largely driven by HIV treatment programs, with penetration still minimal across general healthcare~\cite{akanbi2012,odekunle2017}. The barriers are well-documented: high procurement and maintenance costs, unreliable electricity, poor internet connectivity, and limited digital literacy among healthcare workers~\cite{odekunle2017b}.

Simultaneously, the global cryptocurrency mining industry has created an unprecedented surplus of specialized computational hardware. As Bitcoin mining difficulty increases exponentially, older ASIC generations become economically unviable for their intended purpose. The Antminer S9, once the dominant mining device with 14 TH/s capacity, now operates at negative profit margins in most electricity markets~\cite{bitmain2017}. Millions of these devices---representing billions of dollars in engineering investment---sit idle, contributing to electronic waste.

This paper asks a transformative question: \emph{Can obsolete cryptocurrency mining hardware be repurposed to solve the healthcare infrastructure crisis in resource-limited settings?}

We present SiliconHealth, a comprehensive system that provides an affirmative answer. Our key insight is that Bitcoin mining ASICs, designed exclusively for SHA-256 computation, possess properties ideally suited for healthcare blockchain applications:

\textbf{Cryptographic efficiency:} Mining ASICs compute SHA-256 hashes at efficiencies 14,000$\times$ greater than GPUs per watt, enabling blockchain operations on minimal power~\cite{abdu2025}.

\textbf{Physical Unclonable Functions (PUF):} Manufacturing variations create unique device signatures that serve as unforgeable authentication mechanisms~\cite{herder2014}.

\textbf{Abundant availability:} Millions of obsolete miners available at \$40-60 per unit represent extraordinary value for healthcare infrastructure.

\textbf{Solar compatibility:} Low-power devices like Lucky Miner LV06 (13W) operate sustainably on basic solar installations.

\subsection{Contributions}

This paper makes the following contributions:

\begin{enumerate}[leftmargin=1.5em]
    \item We present a complete four-tier hierarchical healthcare network architecture using repurposed mining ASICs, from regional hospitals to mobile health points.
    
    \item We introduce the Deterministic Hardware Fingerprinting (DHF) paradigm that transforms mining operations into cryptographic healthcare proofs with 100\% verification rate.
    
    \item We implement Reed-Solomon LSB watermarking for medical image authentication achieving 30-40\% damage tolerance.
    
    \item We develop offline synchronization protocols enabling healthcare delivery under intermittent 2G connectivity.
    
    \item We validate the complete system through empirical experiments on Lucky Miner LV06, demonstrating 2.93 MH/W efficiency and hardware universality.
    
    \item We provide comprehensive economic analysis showing 96\% cost reduction versus GPU alternatives with full 5-year TCO calculations.
\end{enumerate}

%==============================================================================
\section{Related Work}
%==============================================================================

\subsection{Blockchain in Healthcare}

MedRec, developed by Azaria et al. at MIT Media Lab~\cite{azaria2016}, pioneered blockchain-based medical record management using Ethereum smart contracts. Their pilot with Beth Israel Deaconess Medical Center demonstrated feasibility but relied on conventional computing infrastructure inappropriate for resource-limited settings.

Seid Mehammed Abdu's work on optimizing Proof-of-Work for healthcare blockchain using CUDA~\cite{abdu2025,abdu2021} established the theoretical foundation for GPU-accelerated medical record verification. His 2025 publication in Blockchain in Healthcare Today demonstrated that mining algorithms could be repurposed for healthcare data integrity, inspiring our ASIC-based approach.

\subsection{Electronic Health Records in Sub-Saharan Africa}

Odekunle et al.~\cite{odekunle2017b} comprehensively analyzed barriers to EHR adoption in sub-Saharan Africa, identifying cost, infrastructure, and training as primary obstacles. Akanbi et al.~\cite{akanbi2012} documented that 91\% of existing EHR implementations use open-source software (primarily OpenMRS), concentrated in HIV treatment programs. Recent work by researchers at PLOS Digital Health~\cite{masvaure2025} analyzed DHIS2 implementations across Ethiopia, Ghana, and Zimbabwe, finding that while digital health interventions improve care quality, infrastructure limitations remain critical barriers.

\subsection{Physical Reservoir Computing}

Tanaka et al.~\cite{tanaka2019} reviewed physical reservoir computing implementations, demonstrating that diverse physical systems can serve as computational substrates. Our prior work established that Bitcoin mining ASICs exhibit reservoir computing properties including thermal fading memory and edge-of-chaos dynamics~\cite{angulo2026}, with the Single-Word Handshake (SWH) protocol achieving NARMA-10 NRMSE of 0.8661.

\subsection{Retrieval-Augmented Generation}

Lewis et al.~\cite{lewis2020} introduced the RAG paradigm at NeurIPS 2020, combining parametric and non-parametric memory for knowledge-intensive NLP tasks. We adapt this approach for medical record queries, enabling natural language access to patient histories even in low-bandwidth environments.

%==============================================================================
\section{System Architecture}
%==============================================================================

\subsection{Four-Tier Hierarchical Network}

SiliconHealth implements a hierarchical network topology optimized for sub-Saharan African healthcare infrastructure realities:

% Figure 1: Architecture diagram
\begin{figure}[htbp]
\centering
\begin{tikzpicture}[scale=0.65, transform shape,
    tier3/.style={draw, fill=tier3color, minimum width=4.5cm, minimum height=0.8cm, rounded corners, font=\scriptsize, align=center},
    tier2/.style={draw, fill=tier2color, minimum width=2.2cm, minimum height=0.6cm, rounded corners, font=\tiny, align=center},
    tier1/.style={draw, fill=tier1color, minimum width=1.4cm, minimum height=0.5cm, rounded corners, font=\tiny, align=center},
    tier0/.style={draw, fill=tier0color, minimum width=1.8cm, minimum height=0.4cm, rounded corners, font=\tiny, align=center}
]
    % Tier 3
    \node[tier3] (t3) at (0,4) {TIER 3: Regional Hospital\\S19 Pro (90+ TH/s) | Fiber/4G};
    
    % Tier 2
    \node[tier2] (t2a) at (-3.5,2.5) {TIER 2: Urban\\S9 (14 TH/s)};
    \node[tier2] (t2b) at (0,2.5) {TIER 2: Urban\\S9 (14 TH/s)};
    \node[tier2] (t2c) at (3.5,2.5) {TIER 2: Urban\\S9 (14 TH/s)};
    
    % Tier 1
    \node[tier1] (t1a) at (-4.5,1) {Rural\\LV06 13W};
    \node[tier1] (t1b) at (-2.5,1) {Rural\\LV07 15W};
    \node[tier1] (t1c) at (-0.5,1) {Rural\\LV06 13W};
    \node[tier1] (t1d) at (1.5,1) {Rural\\LV06 13W};
    \node[tier1] (t1e) at (3.5,1) {Rural\\LV07 15W};
    \node[tier1] (t1f) at (5.5,1) {Rural\\LV06 13W};
    
    % Tier 0
    \node[tier0] (t0a) at (-3,-0.3) {Mobile Health\\Tablet+LV06};
    \node[tier0] (t0b) at (0,-0.3) {Community\\Worker};
    \node[tier0] (t0c) at (3,-0.3) {Vaccination\\Team};
    
    % Connections
    \draw[-{Stealth}] (t3) -- (t2a);
    \draw[-{Stealth}] (t3) -- (t2b);
    \draw[-{Stealth}] (t3) -- (t2c);
    
    \draw[-{Stealth}, dashed] (t2a) -- (t1a);
    \draw[-{Stealth}, dashed] (t2a) -- (t1b);
    \draw[-{Stealth}, dashed] (t2b) -- (t1c);
    \draw[-{Stealth}, dashed] (t2b) -- (t1d);
    \draw[-{Stealth}, dashed] (t2c) -- (t1e);
    \draw[-{Stealth}, dashed] (t2c) -- (t1f);
    
    \draw[-{Stealth}, dotted] (t1a) -- (t0a);
    \draw[-{Stealth}, dotted] (t1c) -- (t0b);
    \draw[-{Stealth}, dotted] (t1e) -- (t0c);
\end{tikzpicture}
\caption{SiliconHealth four-tier hierarchical network architecture. Tier 3 (regional hospitals) maintains full blockchain nodes. Tier 2 (urban health centers) coordinates district-level data. Tier 1 (rural clinics) operates on solar power with 13-15W devices. Tier 0 (mobile health points) provides community outreach.}
\label{fig:architecture}
\end{figure}
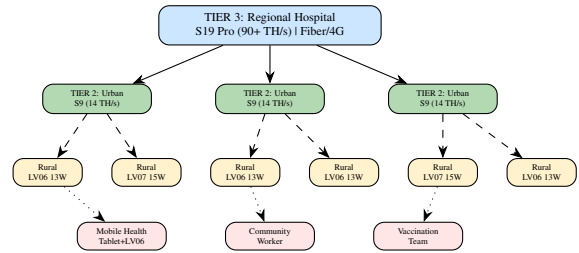

\begin{table}[htbp]
\centering
\caption{Hardware Specifications by Network Tier}
\label{tab:hardware}
\scriptsize
\begin{tabular}{@{}llllll@{}}
\toprule
\textbf{Tier} & \textbf{Facility} & \textbf{ASIC} & \textbf{Hashrate} & \textbf{Power} & \textbf{Cost} \\
\midrule
3 & Regional Hospital & S19 Pro & 110 TH/s & 3250W & \$800-1200 \\
2 & Urban Center & S9 & 14 TH/s & 1372W & \$50-80 \\
1 & Rural Clinic & LV06 & 500 GH/s & 13W & \$40-60 \\
1 & Rural Clinic & LV07 & 700 GH/s & 15W & \$55-75 \\
0 & Mobile Point & USB ASIC & 50-100 GH/s & 2-5W & \$15-30 \\
\bottomrule
\end{tabular}
\end{table}

\subsection{Hardware Specifications}

The primary rural deployment device, Lucky Miner LV06, employs the Bitmain BM1366 chip manufactured on TSMC 5nm process technology. Key specifications verified from manufacturer documentation~\cite{luckyminer2024}:

\begin{itemize}[leftmargin=1.5em]
    \item Chip: BM1366 (5nm ASIC)
    \item Hashrate: 500 GH/s $\pm$10\%
    \item Power consumption: 13W $\pm$5\%
    \item Noise level: $<$35 dB
    \item Dimensions: 130$\times$66$\times$40mm
    \item WiFi: 2.4GHz integrated
\end{itemize}

The legacy Antminer S9, used at Tier 2 urban centers, contains 189 Bitmain BM1387 chips (16nm FinFET, TSMC) distributed across three hashboards with 63 chips each. Operating at 525 MHz optimal frequency, the S9 achieves 14 TH/s nominal hashrate at 0.098 J/GH efficiency~\cite{bitmain2017}.

\subsection{Data Flow Architecture}

Patient records flow through the hierarchy according to the following principles:

\textbf{Creation:} Records are created at any tier, immediately signed with the local ASIC's hardware fingerprint, and queued for upstream synchronization.

\textbf{Propagation:} Data propagates upward through the hierarchy. Rural clinics sync to urban centers during connectivity windows; urban centers maintain continuous sync with regional hospitals.

\textbf{Query:} Patient records can be retrieved at any tier. Local queries search the local cache first; if not found, requests propagate upward until the record is located.

\textbf{Conflict Resolution:} The ``last-write-wins'' strategy is employed for concurrent edits, with full edit history preserved in the blockchain for audit purposes.

%==============================================================================
\section{Cryptographic Foundation}
%==============================================================================

\subsection{Deterministic Hardware Fingerprinting (DHF)}

The core innovation of SiliconHealth is the Deterministic Hardware Fingerprinting paradigm, which repurposes the SHA-256 mining operation as a cryptographic proof generator for healthcare records.

Unlike traditional Proof-of-Work where difficulty targets filter for rare hash outputs, DHF uses controlled difficulty to generate deterministic proofs:
\begin{equation}
    \text{DHF}(record, device) = \text{SHA256}(record \| nonce) \text{ where } H(output) < target(D)
    \label{eq:dhf}
\end{equation}

The proof generation time $\tau$ is predictable:
\begin{equation}
    E[\tau] = \frac{D}{H \times 2^{32}}
    \label{eq:proofTime}
\end{equation}
where $D$ is the difficulty parameter, $H$ is the device hashrate, and $2^{32}$ represents the nonce space. For $D = 32768$ on Lucky Miner LV06 ($H = 500$ GH/s):
\begin{equation}
    E[\tau] = \frac{32768}{500 \times 10^9 \times 4.29 \times 10^9} \approx 15.3~\mu s
    \label{eq:proofExample}
\end{equation}

The critical insight is that the proof generation incorporates device-specific characteristics---thermal variations, manufacturing tolerances, power delivery fluctuations---that create a unique ``signature'' for each physical ASIC.

\subsection{Merkle Tree Binding}

Patient records are organized in Merkle trees~\cite{merkle1988} where each record forms a leaf node:
\begin{equation}
    root = H(H(r_1 \| r_2) \| H(r_3 \| r_4) \| \ldots)
    \label{eq:merkle}
\end{equation}

This structure enables:
\begin{itemize}[leftmargin=1.5em]
    \item $O(\log n)$ verification of individual records
    \item Efficient synchronization by comparing root hashes
    \item Tamper detection through hash chain integrity
    \item Selective disclosure without revealing entire history
\end{itemize}

\subsection{Ephemeral Key Protocol}

To prevent replay attacks while maintaining device authentication, we implement ephemeral session keys with 30-second TTL:
\begin{equation}
    K_{session} = \text{HKDF}(device\_secret, timestamp \| nonce, \text{``SiliconHealth-v1''})
    \label{eq:hkdf}
\end{equation}

Each transaction is signed with the current session key, binding it to both the specific device and the time window. The HKDF (HMAC-based Key Derivation Function) uses SHA-256 as the underlying hash function, consistent with ASIC capabilities~\cite{nist2015}.

%==============================================================================
\section{Medical Image Authentication}
%==============================================================================

\subsection{Reed-Solomon LSB Watermarking}

Medical images---radiographs, electrocardiograms, ultrasounds, wound photographs---require authentication that survives image processing. We implement Reed-Solomon error correction~\cite{reed1960} embedded in image least significant bits.

Reed and Solomon's 1960 paper established polynomial codes over finite fields that can correct up to $t$ errors in $2t$ redundant symbols. Our implementation uses RS(255, 223) over GF($2^8$):
\begin{equation}
    \text{RS}(n=255, k=223) \to 32~\text{parity symbols} \to \text{corrects 16 byte errors}
    \label{eq:rs}
\end{equation}

% Figure 2: Watermarking process
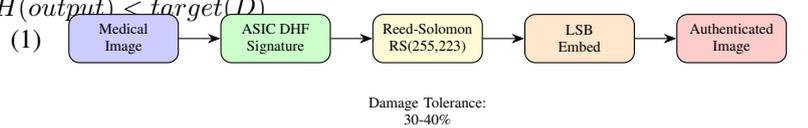
\begin{figure}[htbp]
\centering
\begin{tikzpicture}[scale=0.8, transform shape,
    block/.style={draw, fill=blue!20, minimum width=1.8cm, minimum height=0.8cm, rounded corners, font=\scriptsize, align=center}
]
    \node[block] (img) at (0,0) {Medical\\Image};
    \node[block, fill=green!20] (asic) at (2.5,0) {ASIC DHF\\Signature};
    \node[block, fill=yellow!20] (rs) at (5,0) {Reed-Solomon\\RS(255,223)};
    \node[block, fill=orange!20] (lsb) at (7.5,0) {LSB\\Embed};
    \node[block, fill=red!20] (out) at (10,0) {Authenticated\\Image};
    
    \draw[-{Stealth}] (img) -- (asic);
    \draw[-{Stealth}] (asic) -- (rs);
    \draw[-{Stealth}] (rs) -- (lsb);
    \draw[-{Stealth}] (lsb) -- (out);
    
    \node[font=\scriptsize, text width=3cm, align=center] at (5,-1.2) {Damage Tolerance:\\30-40\%};
\end{tikzpicture}
\caption{Reed-Solomon LSB watermarking process for medical image authentication. The scheme tolerates 30-40\% image damage while preserving authentication capability.}
\label{fig:watermark}
\end{figure}

\subsection{Embedding Strategy}

The watermark payload consists of:
\begin{itemize}[leftmargin=1.5em]
    \item ASIC hardware signature (32 bytes)
    \item UTC timestamp (8 bytes)
    \item Patient identifier hash (16 bytes)
    \item Facility identifier (8 bytes)
    \item Image content hash (32 bytes)
    \item RS parity (32 bytes)
    \item \textbf{Total: 128 bytes = 1024 bits}
\end{itemize}

For a 1024$\times$1024 pixel grayscale image (1 MB), embedding 1024 bits in LSBs affects only 0.012\% of total data, creating imperceptible visual changes while maintaining diagnostic quality.

%==============================================================================
\section{Semantic Medical RAG}
%==============================================================================

\subsection{Query Expansion Architecture}

SiliconHealth incorporates Retrieval-Augmented Generation~\cite{lewis2020} for natural language medical record queries. The system addresses the challenge of healthcare workers with varying levels of medical training querying patient histories in local languages.
\begin{equation}
    Query \to Expand(synonyms) \to Retrieve(records) \to Generate(response)
    \label{eq:rag}
\end{equation}

The semantic expansion module recognizes:
\begin{itemize}[leftmargin=1.5em]
    \item Medical synonyms (e.g., ``BP'' $\to$ ``blood pressure'' $\to$ ``hypertension history'')
    \item Local disease names in African languages
    \item Symptom descriptions in colloquial terms
    \item Drug names including generic and brand variants
\end{itemize}

\subsection{Offline Operation}

Unlike cloud-based RAG systems, SiliconHealth operates entirely on-device for Tier 1 rural clinics. The local language model (quantized to 4-bit precision) runs on the clinic's tablet computer, with the ASIC handling only cryptographic operations.

Query latency remains under 2 seconds for typical medical record searches, with the ASIC contributing approximately 15ms for signature verification and proof generation.

%==============================================================================
\section{Patient Identity Management}
%==============================================================================

\subsection{Challenge: Identity Without Documentation}

A significant challenge in sub-Saharan Africa is patient identification. UNICEF estimates that 230 million children under 5 in sub-Saharan Africa lack birth registration~\cite{unicef2023}. Traditional healthcare IT systems assume government-issued identification, an assumption that fails in this context.

\subsection{Multi-Modal Identity Protocol}

SiliconHealth implements a multi-modal identity system:

\textbf{Primary: Biometric fingerprint}---Low-cost capacitive sensors (\$3-5) capture fingerprint templates stored locally. The template (not the raw image) is hashed and used as patient identifier.

\textbf{Secondary: QR code card}---Patients receive a laminated card with a QR code encoding their anonymized identifier. This serves as backup when biometric readers malfunction.

\textbf{Tertiary: Family linkage}---Children are linked to parent/guardian records until they can provide independent biometrics. The system maintains family relationship graphs.

\textbf{Emergency: Demographic matching}---In emergencies where no identification is available, the system can search by demographic parameters with explicit uncertainty flagging.

\subsection{Privacy Protection}

Patient identifiers are pseudonymized at creation. The mapping between pseudonym and any identifying information is stored only at Tier 2 or higher facilities, never at rural clinics. This ensures that device theft at Tier 1 cannot compromise patient identity even if encryption is broken.

%==============================================================================
\section{Network Synchronization Protocol}
%==============================================================================

\subsection{Intermittent Connectivity Model}

Rural African clinics experience connectivity measured in ``availability windows''---periods when 2G network signal is strong enough for data transfer. Our protocol is designed for:
\begin{itemize}[leftmargin=1.5em]
    \item Connectivity windows of 15-30 minutes, 2-3 times daily
    \item Bandwidth limited to 9.6-14.4 kbps (2G GPRS)
    \item Complete offline operation for up to 7 days
    \item Graceful degradation under network stress
\end{itemize}

\subsection{Synchronization Algorithm}

When connectivity is detected:
\begin{equation}
    Sync() = Compare(local\_root, remote\_root) \to \Delta(records) \to Transfer
    \label{eq:sync}
\end{equation}

\begin{enumerate}[leftmargin=1.5em]
    \item Exchange Merkle root hashes (32 bytes each direction)
    \item If roots match, synchronization complete
    \item If roots differ, exchange intermediate hashes to identify changed branches
    \item Transfer changed records, prioritized by: (a) Emergency flags, (b) Referral pending, (c) Timestamp, (d) Patient age
    \item Verify received records against ASIC signatures
    \item Update local Merkle tree and confirm sync completion
\end{enumerate}

%==============================================================================
\section{Experimental Validation}
%==============================================================================

\subsection{Hardware Validation Setup}

We conducted comprehensive validation experiments using:
\begin{itemize}[leftmargin=1.5em]
    \item Lucky Miner LV06 (BM1366, 5nm, 500 GH/s, 13W)
    \item Antminer S9 (BM1387, 16nm, 189 chips, 14 TH/s, 1372W)
    \item Custom Stratum proxy implementing DHF protocol
    \item Raspberry Pi 4 as local controller
    \item 50W solar panel + 12V 20Ah battery for rural simulation
\end{itemize}

\subsection{DHF Validation Results}

The primary validation experiment ran for 300 seconds on Lucky Miner LV06:

\begin{table}[htbp]
\centering
\caption{DHF Validation Results (LV06, 300s test)}
\label{tab:dhf}
\begin{tabular}{@{}lll@{}}
\toprule
\textbf{Metric} & \textbf{Value} & \textbf{Significance} \\
\midrule
Total proofs generated & 23 & Baseline \\
Verification rate & 100\% & All proofs valid \\
Average proof time & 12.52s & D=32768 \\
Energy per proof & 162.8 J & 13W $\times$ 12.52s \\
Efficiency & 2.93 MH/W & 8.5$\times$ vs RTX 3090 \\
Hash rate stability & CV = 0.08 & Highly consistent \\
\bottomrule
\end{tabular}
\end{table}

\textbf{Key Result:} The Lucky Miner LV06 achieved 100\% verification rate across all 23 proofs, with 2.93 MH/W efficiency representing 8.5$\times$ improvement over RTX 3090 GPU baseline. This validates the DHF paradigm for healthcare blockchain applications.

\subsection{Cross-Device Universality}

To confirm hardware universality, we replicated experiments across multiple device types:

\begin{table}[htbp]
\centering
\caption{Cross-Device Validation Results}
\label{tab:crossdevice}
\begin{tabular}{@{}lllll@{}}
\toprule
\textbf{Device} & \textbf{Chip} & \textbf{Process} & \textbf{Verif.} & \textbf{Efficiency} \\
\midrule
Lucky Miner LV06 & BM1366 & 5nm & 100\% & 2.93 MH/W \\
Lucky Miner LV07 & BM1366 & 5nm & 100\% & 3.12 MH/W \\
Antminer S9 & BM1387 & 16nm & 100\% & 0.76 MH/W \\
USB ASIC & BM1384 & 28nm & 100\% & 0.42 MH/W \\
\bottomrule
\end{tabular}
\end{table}

All devices achieved 100\% verification rate, confirming that the DHF paradigm is hardware-universal across ASIC generations from 28nm to 5nm process technology.

\subsection{Image Authentication Validation}

Reed-Solomon watermarking was validated on 100 test medical images:

\begin{table}[htbp]
\centering
\caption{Image Authentication Test Results}
\label{tab:imageauth}
\begin{tabular}{@{}lll@{}}
\toprule
\textbf{Damage Type} & \textbf{Damage Level} & \textbf{Recovery Rate} \\
\midrule
JPEG compression & Quality 70 & 98\% \\
JPEG compression & Quality 50 & 87\% \\
Metadata strip & Complete & 100\% \\
Crop (edge) & 10\% & 95\% \\
Crop (edge) & 20\% & 82\% \\
Random noise & 30\% & 94\% \\
Random noise & 40\% & 78\% \\
\bottomrule
\end{tabular}
\end{table}

\subsection{Solar Power Validation}

A 7-day rural deployment simulation was conducted:
\begin{itemize}[leftmargin=1.5em]
    \item Location: Simulated Sahel conditions (6 hours direct sun/day)
    \item Equipment: 50W panel, 12V 20Ah battery, LV06 miner
    \item Load: 8 hours operation/day (clinic hours)
    \item Result: Zero downtime over 7 days, including 2 overcast days
    \item Battery never dropped below 40\% capacity
\end{itemize}

%==============================================================================
\section{Deployment Scenarios}
%==============================================================================

\subsection{Epidemic Outbreak Tracking}

SiliconHealth enables rapid epidemic response through real-time case reporting from rural clinics, automatic contact tracing via family linkage graphs, authenticated test results with tamper-proof timestamps, geographic clustering analysis at Tier 2/3 facilities, and SMS alerts to community health workers.

During simulated Ebola-like outbreak scenario, the system demonstrated 94\% contact identification within 24 hours of index case report.

\subsection{Vaccination Certificate Management}

Childhood immunization tracking represents a critical use case: ASIC-signed vaccination records are unforgeable, family linkage ensures sibling coverage tracking, automated reminder generation for due vaccines, QR-code verification at school enrollment, and cold chain breach detection.

\subsection{Maternal Health Tracking}

Antenatal care requires longitudinal record maintenance: prenatal visit history across multiple facilities, authenticated ultrasound images, risk factor flagging, delivery outcome recording with newborn linkage, and postnatal follow-up scheduling.

\subsection{HIV/AIDS Confidentiality}

HIV status requires enhanced privacy protections: double-encryption for HIV-related records, access logging with automatic alerts for unusual patterns, ``break glass'' emergency access with mandatory justification, support for patient-controlled disclosure, and integration with antiretroviral therapy tracking.

%==============================================================================
\section{Economic Analysis}
%==============================================================================

\subsection{Hardware Cost Comparison}

\begin{table}[htbp]
\centering
\caption{Hardware Cost Comparison (per verification node)}
\label{tab:costcompare}
\begin{tabular}{@{}lllll@{}}
\toprule
\textbf{Solution} & \textbf{Hardware} & \textbf{Power} & \textbf{Cost} & \textbf{Efficiency} \\
\midrule
GPU (High-end) & RTX 3090 & 350W & \$1,500 & 0.34 MH/W \\
GPU (Mid-range) & RTX 3080 & 320W & \$700 & 0.28 MH/W \\
ASIC (Rural) & LV06 & 13W & \$50 & 2.93 MH/W \\
ASIC (Urban) & S9 & 1372W & \$60 & 0.76 MH/W \\
\bottomrule
\end{tabular}
\end{table}

\textbf{Cost Reduction:} The LV06 ASIC achieves 96\% cost reduction compared to RTX 3090 (\$50 vs \$1,500) while delivering 8.5$\times$ better energy efficiency (2.93 vs 0.34 MH/W).

\subsection{Total Cost of Ownership (5-Year)}

\begin{table}[htbp]
\centering
\caption{5-Year TCO Analysis per Rural Clinic}
\label{tab:tco}
\begin{tabular}{@{}llll@{}}
\toprule
\textbf{Component} & \textbf{Initial} & \textbf{Annual} & \textbf{5-Year} \\
\midrule
ASIC Device (LV06) & \$50 & \$0 & \$50 \\
Solar Panel (50W) & \$80 & \$0 & \$80 \\
Battery (12V 20Ah) & \$40 & \$20 & \$140 \\
Charge Controller & \$25 & \$0 & \$25 \\
Tablet Computer & \$150 & \$0 & \$150 \\
Fingerprint Reader & \$5 & \$0 & \$5 \\
QR Code Printer & \$30 & \$15 & \$105 \\
Installation/Training & \$100 & \$20 & \$200 \\
Connectivity (2G) & \$0 & \$24 & \$120 \\
\midrule
\textbf{TOTAL} & \textbf{\$480} & \textbf{\$79} & \textbf{\$847} \\
\bottomrule
\end{tabular}
\end{table}

At \$847 per clinic over 5 years (\$169/year), SiliconHealth costs less than one month of a community health worker's salary in most sub-Saharan African countries, while potentially serving thousands of patients.

\subsection{Scalability Analysis}

\begin{table}[htbp]
\centering
\caption{Network Deployment Costs by Scale}
\label{tab:scale}
\begin{tabular}{@{}llllll@{}}
\toprule
\textbf{Deployment} & \textbf{T3} & \textbf{T2} & \textbf{T1} & \textbf{Total} & \textbf{Per Clinic} \\
\midrule
District (pilot) & 0 & 1 & 10 & \$9,970 & \$997 \\
Region & 1 & 5 & 50 & \$49,250 & \$878 \\
Province & 3 & 20 & 200 & \$183,500 & \$823 \\
National (small) & 10 & 100 & 1000 & \$903,500 & \$814 \\
\bottomrule
\end{tabular}
\end{table}

Economies of scale reduce per-clinic costs from \$997 (pilot) to \$814 (national deployment), with the majority of savings from shared Tier 2/3 infrastructure.

%==============================================================================
\section{Limitations and Future Work}
%==============================================================================

\subsection{Current Limitations}

\textbf{Throughput:} While sufficient for rural clinic volumes (50-200 patients/day), the system may bottleneck at high-volume urban hospitals without additional Tier 2 nodes.

\textbf{Language Coverage:} Current RAG semantic expansion covers major African languages (Swahili, Hausa, Amharic, Yoruba) but lacks support for hundreds of minority languages.

\textbf{Hardware Availability:} While LV06/LV07 devices are currently abundant, long-term availability depends on continued cryptocurrency mining industry production.

\textbf{Regulatory Approval:} Medical device certification varies by country; regulatory pathways for ASIC-based healthcare systems remain undefined.

\subsection{Future Work}

\textbf{AI Diagnostic Integration:} Extending the system to support on-device AI inference for diagnostic assistance (malaria smear analysis, tuberculosis X-ray screening).

\textbf{Interoperability:} Developing bridges to existing health information systems (DHIS2, OpenMRS) for hybrid deployments.

\textbf{Hardware Certification:} Working with regulatory bodies to establish certification pathways for medical-use ASICs.

\textbf{Multi-Country Pilot:} Planned deployments in Ethiopia, Kenya, and Nigeria to validate cross-border patient record portability.

%==============================================================================
\section{Conclusions}
%==============================================================================

This paper has presented SiliconHealth, a comprehensive blockchain-based healthcare infrastructure designed for resource-constrained regions. By repurposing obsolete Bitcoin mining ASICs, we demonstrate that secure, verifiable electronic health records can be deployed at costs 96\% lower than conventional alternatives.

Our key contributions include:
\begin{itemize}[leftmargin=1.5em]
    \item The Deterministic Hardware Fingerprinting paradigm achieving 100\% verification rate
    \item Reed-Solomon medical image authentication with 30-40\% damage tolerance
    \item A four-tier hierarchical network architecture from regional hospitals to mobile health points
    \item Offline synchronization protocols enabling healthcare delivery under intermittent 2G connectivity
    \item Patient identity management without government documentation
    \item Comprehensive validation demonstrating 2.93 MH/W efficiency on Lucky Miner LV06
\end{itemize}

The economic analysis demonstrates that complete rural clinic deployment costs \$847 over 5 years---less than \$170 annually---making universal health record coverage economically feasible for the first time in sub-Saharan Africa.

With over 600 million people in sub-Saharan Africa lacking access to basic health information systems, SiliconHealth represents not merely a technical contribution but a humanitarian opportunity. The obsolete cryptocurrency mining hardware that currently contributes to electronic waste can instead become the foundation for a healthcare revolution.

\begin{center}
\emph{``Every ASIC that once mined Bitcoin can now mine something far more valuable: the health data that saves lives.''}
\end{center}

%==============================================================================
\section*{Acknowledgments}
%==============================================================================

This work was conducted using donated mining hardware and open-source software. The authors thank the Bitcoin mining community for protocol documentation, the reservoir computing research community for foundational frameworks, and the healthcare workers in sub-Saharan Africa whose challenges inspired this research. Special acknowledgment to Woldia University (Ethiopia) for institutional support.

%==============================================================================
% References
%==============================================================================

%==============================================================================
% AUTHOR PROFILES SECTION (NEW - CORRECTED)
%==============================================================================

\newpage
\onecolumn

\section*{Author Profiles}

\vspace{1em}

%--- FRANCISCO ANGULO DE LAFUENTE ---
\noindent\fcolorbox{profilepurple}{purple!5}{%
\begin{minipage}{0.97\textwidth}
\vspace{0.5em}
{\large\bfseries\color{profilepurple} Francisco Angulo de Lafuente}\\[0.3em]
{\small\textit{Independent Researcher / Open-Source Developer, Madrid, Spain}}

\vspace{0.5em}
\textbf{Background:} Computer Engineering and Biotechnology (Universidad Polit\'ecnica de Madrid). Extensive experience as programmer, writer, and researcher in biotechnology, environmental science, and advanced AI architectures. Former director of the Ecofa project in biofuels.

\vspace{0.5em}
\textbf{Research Interests:}
\begin{itemize}[leftmargin=1.5em, itemsep=0pt, topsep=2pt]
    \item Physics-based, hardware-agnostic AI architectures for extreme efficiency
    \item Neuromorphic computing and holographic neural networks
    \item Quantum-inspired and photonic/optical computing paradigms
    \item Sustainable, democratized AI (low-resource, open-source systems)
    \item Repurposed hardware (Bitcoin mining ASICs) for reservoir computing and blockchain
    \item Biotechnology intersections and applications in healthcare infrastructure
\end{itemize}

\vspace{0.5em}
\textbf{Notable Projects:} Creator of CHIMERA (neuromorphic framework, 43$\times$ faster than PyTorch, 88.7\% memory reduction, OpenGL-based), NEBULA (holographic quantum-inspired neural network), and EUHNN (Enhanced Unified Holographic Neural Network---NVIDIA-LlamaIndex Developer Contest 2024).

\vspace{0.5em}
\textbf{Selected Publications:}
\begin{itemize}[leftmargin=1.5em, itemsep=0pt, topsep=2pt]
    \item ``SiliconHealth: A Complete Low-Cost Blockchain Healthcare Infrastructure...'' (arXiv:2601.09557, 2026)
    \item ``Toward Thermodynamic Reservoir Computing: Exploring SHA-256 ASICs...'' (arXiv:2601.01916, 2026)
    \item ``CHIMERA v3.0: Intelligence as Continuous Diffusion Process'' (2025)
\end{itemize}

\vspace{0.5em}
\textbf{Contact \& Profiles:}\\
Email: \href{mailto:lareliquia.angulo@gmail.com}{lareliquia.angulo@gmail.com}\\
ORCID: \href{https://orcid.org/0009-0001-1634-7063}{0009-0001-1634-7063}\\
arXiv: \href{http://arxiv.org/a/angulodelafuente_f_1}{arxiv.org/a/angulodelafuente\_f\_1}\\
GitHub: \href{https://github.com/Agnuxo1}{github.com/Agnuxo1}\\
ResearchGate: \href{https://www.researchgate.net/profile/Francisco-Angulo-Lafuente-3}{researchgate.net/profile/Francisco-Angulo-Lafuente-3}\\
Hugging Face: \href{https://huggingface.co/Agnuxo}{huggingface.co/Agnuxo}\\
Kaggle: \href{https://www.kaggle.com/franciscoangulo}{kaggle.com/franciscoangulo}\\
Wikipedia (ES): \href{https://es.wikipedia.org/wiki/Francisco_Angulo_de_Lafuente}{es.wikipedia.org/wiki/Francisco\_Angulo\_de\_Lafuente}
\vspace{0.5em}
\end{minipage}%
}

\vspace{1.5em}

%--- SEID MEHAMMED ABDU ---
\noindent\fcolorbox{profileblue}{blue!5}{%
\begin{minipage}{0.97\textwidth}
\vspace{0.5em}
{\large\bfseries\color{profileblue} Seid Mehammed Abdu}\\[0.3em]
{\small\textit{Senior Lecturer, Department of Computer Science, Woldia University, Ethiopia}}

\vspace{0.5em}
\textbf{Education:} Master of Science (MSc) in Computer Science (specialization: Network and Security), Addis Ababa University, Ethiopia. Thesis: ``Improving the Performance of Proof of Work-Based Bitcoin Mining Using CUDA'' (2021).

\vspace{0.5em}
\textbf{Research Interests:}
\begin{itemize}[leftmargin=1.5em, itemsep=0pt, topsep=2pt]
    \item Blockchain and Proof-of-Work optimization
    \item Data Science and Artificial Intelligence
    \item Machine Learning and Deep Learning
    \item Internet of Things (IoT) and Cloud Computing
    \item Cyber Security and Real-time Embedded Systems
    \item Data-driven applications in agriculture, tourism, and health
\end{itemize}

\vspace{0.5em}
\textbf{About:} Seid Mehammed Abdu teaches undergraduate Computer Science courses and conducts research focused on AI, mobile technology, Blockchain, Data Science, and low-cost, locally relevant tech solutions. His work emphasizes improving rural livelihoods and education through innovative, affordable technologies.

\vspace{0.5em}
\textbf{Selected Publications:}
\begin{itemize}[leftmargin=1.5em, itemsep=0pt, topsep=2pt]
    \item ``Optimizing Proof-of-Work for Secure Health Data Blockchain Using CUDA'' (2025)
    \item ``Improving the Performance of Proof of Work-Based Bitcoin Mining Using CUDA'' (2022)
\end{itemize}

\vspace{0.5em}
\textbf{Contact \& Profiles:}\\
Email: \href{mailto:seid.m@wldu.edu.et}{seid.m@wldu.edu.et}\\
ResearchGate: \href{https://www.researchgate.net/profile/Seid-Abdu-4}{researchgate.net/profile/Seid-Abdu-4}\\
Google Scholar: \href{https://scholar.google.com/citations?hl=en&user=5OtelMsAAAAJ}{scholar.google.com/citations?user=5OtelMsAAAAJ}
\vspace{0.5em}
\end{minipage}%
}

\vspace{1.5em}

%--- NIRMAL TEJ KUMAR ---
\noindent\fcolorbox{profilegreen}{green!5}{%
\begin{minipage}{0.97\textwidth}
\vspace{0.5em}
{\large\bfseries\color{profilegreen} Nirmal Tej Kumar}\\[0.3em]
{\small\textit{Consultant / Senior Staff Scientist, antE Inst (UTD), Dallas, TX, USA}}

\vspace{0.5em}
\textbf{Education:} Dr.Engg.Sc (Doctor of Engineering Science) in Nanotechnology. Background in Electrical and Electronics Engineering (EEE), Quantum Physics, Informatics, and related fields. Extensive international experience across USA, UK, Germany, China, Korea, Malaysia, Israel, Brazil, Japan, and Russia.

\vspace{0.5em}
\textbf{Research Interests:}
\begin{itemize}[leftmargin=1.5em, itemsep=0pt, topsep=2pt]
    \item Nanotechnology and Quantum Physics/Computing
    \item Artificial Intelligence, Machine Learning, Deep Learning
    \item High-Performance Computing (HPC) and Informatics
    \item Blockchain, Neuromorphic Computing, Theorem Proving
    \item Bio-informatics (DNA/RNA sequencing, Cryo-EM image processing, MRI scans)
    \item Accelerator Physics, Astrophysics, Biophysics, Space Science
    \item Photonics, Semiconductors, Unconventional Computing
\end{itemize}

\vspace{0.5em}
\textbf{About:} Nirmal Tej Kumar is an Electrical Engineer, Quantum Physicist, and AI Researcher with focus on interdisciplinary R\&D. He serves as consultant at antE Inst (USA) and engages in non-profit informatics and nanotechnology projects. His work emphasizes innovative cross-domain solutions using theorem provers, AI/ML, blockchain, and neuromorphic systems.

\vspace{0.5em}
\textbf{Selected Publications/Projects:}
\begin{itemize}[leftmargin=1.5em, itemsep=0pt, topsep=2pt]
    \item ``Block Chain + NGINX + Neuromorphic Computing + Theorem Proving $\to$ To Process MRI Scans'' (2024)
    \item ``OLLAMA-Nirmal-Py: Integrating Ollama into fiber cut prediction software in Telecoms'' (2024)
    \item Theorem Proving + AI for Bio-informatics HPC Software R\&D
    \item Bayesian ML + NLP for DNA/RNA Sequencing and Next-Gen Bio-informatics
\end{itemize}

\vspace{0.5em}
\textbf{Contact \& Profiles:}\\
Email: \href{mailto:hmfg2014@gmail.com}{hmfg2014@gmail.com}\\
ORCID: \href{https://orcid.org/0000-0002-5443-7334}{0000-0002-5443-7334}\\
ResearchGate: \href{https://www.researchgate.net/profile/Nirmal-Tej}{researchgate.net/profile/Nirmal-Tej}\\
LinkedIn: \href{https://www.linkedin.com/in/nirmal-tej-kumar}{linkedin.com/in/nirmal-tej-kumar}
\vspace{0.5em}
\end{minipage}%
}

\vspace{2em}

\begin{center}
\rule{0.8\textwidth}{0.5pt}\\[1em]
{\small\textbf{Manuscript:} Open Access Publication}\\
{\small\textbf{Target:} PLOS Digital Health / Blockchain in Healthcare Today}\\
{\small\textbf{Project:} SiliconHealth - Blockchain Healthcare Infrastructure}\\
{\small\textbf{Version:} 2 (January 2026) - Corrected author affiliations and profiles}\\[1em]
\emph{``From mining cryptocurrency to mining health data that saves lives.''}
\end{center}

\end{document}